\documentclass[manuscript,screen,accept]{acmart}
\usepackage{booktabs} 
\usepackage{multirow}
\usepackage{graphicx}
\usepackage{color}
\usepackage{url}
\usepackage{footnote}
\makesavenoteenv{tabular}
\makesavenoteenv{table}
\usepackage{hyperref}
\usepackage{amsmath}   
\usepackage{array}
\usepackage{{inputenc}}
\usepackage{balance}
\usepackage{amsmath}
\usepackage{amsfonts}
\usepackage{enumitem}
\usepackage{ragged2e}
\usepackage{longtable}
\usepackage{tabularx}
\usepackage{subcaption}
\usepackage{array}

\usepackage{amsmath,amsfonts,bm}

\def\eqref#1{equation~\ref{#1}}

\def\1{\mathbf{1}}

\def\va{{\mathbf{a}}}

\def\vh{{\mathbf{h}}}

\def\vr{{\mathbf{r}}}
\def\vs{{\mathbf{s}}}

\def\vv{{\mathbf{v}}}

\def\vx{{\mathbf{x}}}
\def\vy{{\mathbf{y}}}
\def\vz{{\mathbf{z}}}

\def\mA{{\mathbf{A}}}

\def\mK{{\mathbf{K}}}

\def\mP{{\mathbf{P}}}
\def\mQ{{\mathbf{Q}}}

\def\mV{{\mathbf{V}}}
\def\mW{{\mathbf{W}}}
\def\mX{{\mathbf{X}}}

\DeclareMathAlphabet{\mathsfit}{\encodingdefault}{\sfdefault}{m}{sl}
\SetMathAlphabet{\mathsfit}{bold}{\encodingdefault}{\sfdefault}{bx}{n}

\def\gE{{\mathcal{E}}}

\def\gG{{\mathcal{G}}}

\def\gN{{\mathcal{N}}}

\def\gP{{\mathcal{P}}}

\def\gV{{\mathcal{V}}}

\def\sN{{\mathbb{N}}}

\def\sP{{\mathbb{P}}}

\def\sR{{\mathbb{R}}}
\def\sS{{\mathbb{S}}}

\newcommand{\eat}[1]{}

\acmJournal{TOIS}

\setcopyright{rightsretained}
\newcommand{\ssgrec}{S$^2$GRec}

\begin{document}
\title{Self-supervised Graph-based Point-of-interest Recommendation}

\author{Yang Li}
\affiliation{%
  \institution{The University of Queensland}
  \city{Brisbane}
  \state{Queensland}
  \country{Australia}
  \postcode{4068}
}
\email{yang.li@uq.edu.au}

\author{Tong Chen}
\affiliation{%
  \institution{The University of Queensland}
  \city{Brisbane}
  \state{Queensland}
  \country{Australia}
  \postcode{4068}
}
\email{tong.chen@uq.edu.au	}

\author{Peng-Fei Zhang}
\affiliation{%
  \institution{The University of Queensland}
  \city{Brisbane}
  \state{Queensland}
  \country{Australia}
  \postcode{4068}
}
\email{mima.zpf@gmail.com}

\author{Zi Huang}
\affiliation{%
  \institution{The University of Queensland}
  \city{Brisbane}
  \state{Queensland}
  \country{Australia}
  \postcode{4068}
}
\email{huang@itee.uq.edu.au}

\author{Hongzhi Yin}
\affiliation{%
  \institution{The University of Queensland}
  \city{Brisbane}
  \state{Queensland}
  \country{Australia}
  \postcode{4068}
}
\email{h.yin1@uq.edu.au}

\begin{abstract}
  The exponential growth of Location-based Social Networks (LBSNs) has greatly stimulated the demand for precise location-based recommendation services. Next Point-of-Interest (POI) recommendation, which aims to provide personalised POI suggestions for users based on their visiting histories, has become the prominent component in location-based e-commerce. Recent POI recommenders mainly employ self-attention mechanism or graph neural networks to model the complex high-order POI-wise interactions. However, most of them are merely trained on the historical check-in data in standard supervised learning manner, which fail to fully explore each user's multi-faceted preferences, and suffer from data scarcity and long-tailed POI distribution, resulting in sub-optimal performance. To this end, we propose a \underline{S}elf-\underline{s}upervised \underline{G}raph-enhanced POI \underline{Rec}ommender (S$^2$GRec) for next POI recommendation. In particular, we devise a novel Graph-enhanced Self-attentive layer to incorporate the collaborative signals from both global transition graph and and local trajectory graphs to uncover the transitional dependencies among POIs and capture a user's temporal interests. In order to counteract the scarcity and incompleteness of POI check-ins, we propose a novel self-supervised learning paradigm in \ssgrec, where the trajectory representations are contrastively learned from two augmented views on geolocations and temporal transitions. Extensive experiments are conducted on three real-world LBSN datasets, demonstrating the effectiveness of our model against state-of-the-art methods.
\end{abstract}

\maketitle
\section{Introduction}
The rapid growth of Location-Based Social Networks (LBSNs), such as Yelp and Foursquare, has significantly raised attention on the studies of location-based recommendation in recent years. Among various location-based services, next Point-of-Interest (POI) recommendation is the most prominent one since it can effectively assist service providers to comprehensively understand user movement patterns for accurate customised promotions. Meanwhile, the next POI recommenders also provide users with tailored trip plans, helping them to make decisions on the next move according to their own travel history.

Previous next POI recommenders have been evolving from pure temporal sequential transitions by first-order Markov-chain \cite{ChengYLK13} and tensor factorisation \cite{zhao2016stellar,HeLLSC16,QianLNY19}, to recurrent neural networks (RNNs) \cite{LiuWWT16,ManotumruksaMO18,ZhaoZLXLZSZ19} and self-attention mechanism \cite{LianWG0C20,LuoLL21} for long-term and short-term preference modelling. In addition to the exploitation of sequential regularities, the RNN-based and self-attentive methods additionally explore the spatiotemporal effects of geographical distances and time intervals between successive check-ins. However, the transition patterns among POIs learned by these methods are limited to a \textbf{local} view, where only the co-occurrences of POIs within each independent check-in sequence are captured. Intuitively, similar users sharing common preferences can be leveraged as a reference for recommending next POI to a certain user. By converting check-in trajectories into graphs, recent graph neural network (GNN)-based approaches \cite{LimHNWGWV20,LiCLYH21} are able to take advantage of \textbf{global} spatial and temporal factors by taking account of the correlated neighbour check-in sequences for recommendation. However, the traditional GNNs, such as graph attention networks \cite{VelickovicCCRLB18} and graph convolutional networks \cite{KipfW17}, suffer from over-smoothing problem when stacking more than 3 layers, which may fail to capture the long-range POI dependencies. In \cite{LiCLYH21}, we propose a new GNN-based POI recommender that incorporates the category-wise correlation modelling to alleviate the negative impacts from the sparse observed POI-POI interactions. However, SGRec still faces the challenges from the following perspectives:
\begin{itemize}
    \item Missing check-in data: prior work purely rely on the observed check-in records for next POI recommendation, where some possible check-ins between each successive check-in pairs are neglected since it is unlikely for a user to check in on each visit. This results in incomplete check-in trajectories, which undermines the capability of the existing work on POI correlation modelling.
    \item Noisy check-in data: there exist many obvious unreal check-in records (e.g., two consecutive check-ins are far apart in distance but occurred in a short time period) and incorrect location ID (e.g., a POI near the actual visiting POI) in raw dataset due to the unstable GPS signal and mobile network signal, which impede the existing models to reveal the actual POI transition correlations, resulting in sub-optimal performance.
\end{itemize}

\begin{figure}[t]
\includegraphics[width=\textwidth]{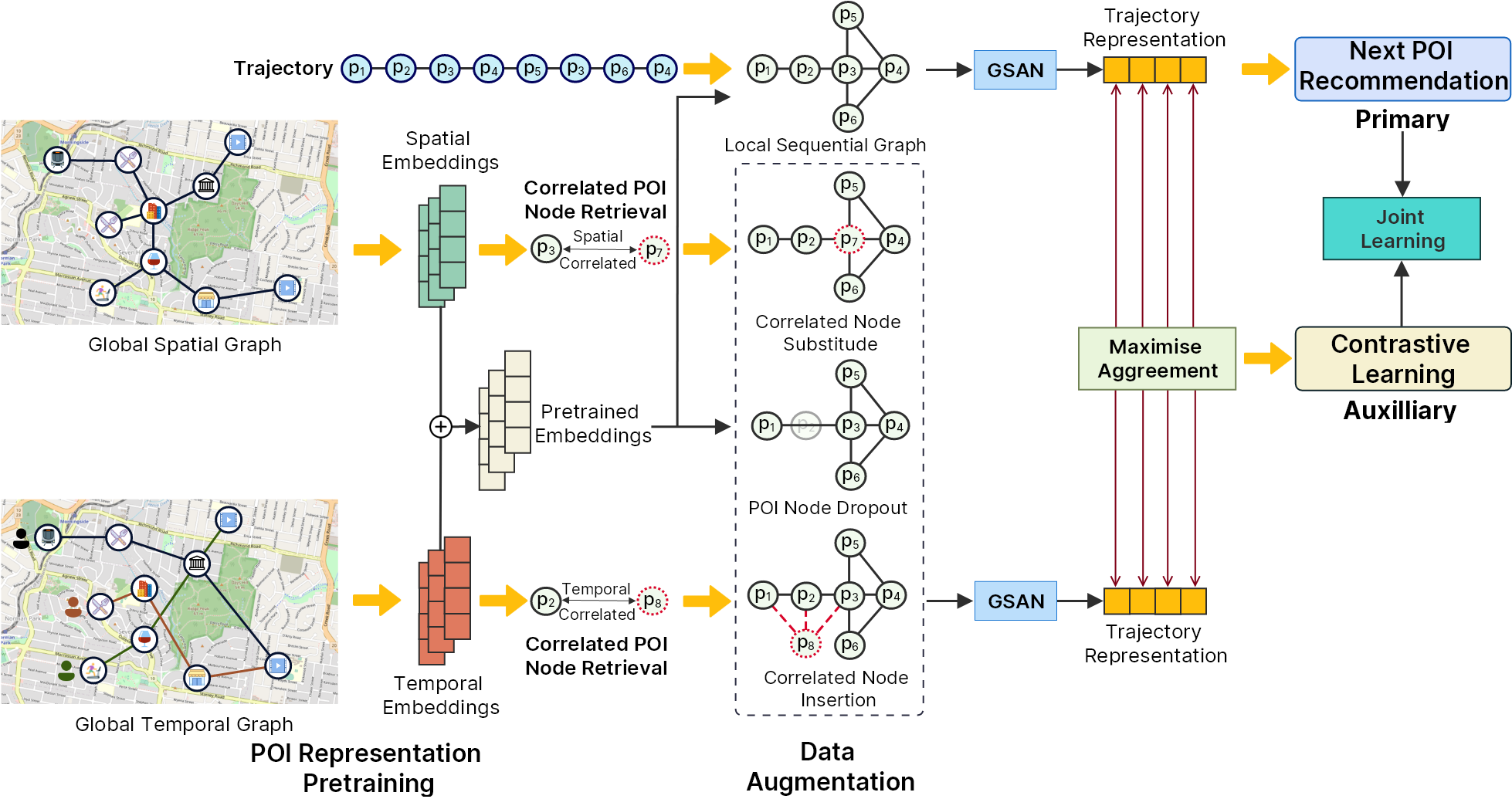}
\caption{A overview of the proposed \ssgrec\ framework.}
\label{fig:model_overview}
\end{figure}

To address the aforementioned limitations and challenges, we make substantial technical advancements upon SGRec \cite{LiCLYH21}, and propose a \textbf{S}elf-\textbf{S}upervised \textbf{G}raph-enhanced POI \textbf{Rec}ommendation framework (\ssgrec). Figure \ref{fig:model_overview} illustrates the overall workflow of \ssgrec. It first derives POI representations from two constructed global graphs in the pretraining stage, making each POI carries the knowledge of global spatial and temporal information. Then, to address the long-range dependency modelling limitations of prior GNN-based approaches and take account of global spatial and temporal factors for recommendation, we propose a novel \textbf{G}raph-enhanced \textbf{S}elf-\textbf{A}ttention \textbf{N}etwork (GSAN) layer as the user preference encoder, which makes full use of the long-short-range dependency modelling capability offered by the self-attention mechanism, and further aggregates the collaborative signals from the global graphs to obtain the comprehensive trajectory representations. Finally, we address the two above-mentioned challenges through an integration of self-supervision signals. Concretely, three carefully-designed augmentation operators are devised, which are performed on the local trajectory graphs to mimic the real-life human check-in behaviours by studying the underlying semantics among POIs, yielding different views of each training instance (i.e., the augmented trajectory graphs). An auxiliary self-supervision task of optimising the consistency between the views of the same instance as well as discriminating it from other training samples is introduced to boost the robustness of \ssgrec\ to the data sparsity and noise.

Our main contributions of this paper are summarised in the followings:
\begin{itemize}
    \item We propose a novel Self-supervised Graph-enhanced POI Recommender (\ssgrec) based on self-attention network to model the complex POI-POI relationships and make fully effective use of global spatial and temporal information for next POI recommendation.
    \item We propose an innovative graph-enhanced self-attentive user preference encoder, GSAN, which adequately captures both of the complex high-order underlying POI-wise and category-wise dependencies from various graph signals, while maintains the sequential properties within the constructed trajectory graphs. 
    \item We introduce a novel auxiliary self-supervision task that complements \ssgrec’s capability of learning user behaviours from sparse and noisy data by effective data augmentation with semantically-correlated POIs. 
    \item Extensive experiments are conducted on three real-world datasets demonstrating the efficacy of our framework against several state-of-the-art next POI recommenders, and a detailed ablation study is launched on the effectiveness of each proposed component.
\end{itemize}

\section{Related Work}
In this section, we highlight the most recent work in next POI recommendation and contrastive self-supervised learning in recommender systems, which are the most relevant to our work.
\subsection{Next POI Recommendation}
Different from the conventional POI recommendation that mainly captures user-POI relationships and geographical impacts from users' check-in activities, next POI recommendation is more challenging since it needs to pay additional attention to the sequential patterns for the user's next move prediction. Intuitively, the chronologically check-in records carry strong spatio-temporal characteristics. Thus, various studies have been conducted to model the sequential dependencies and spatial relationships among POIs for next POI recommendation. In general, the existing next POI recommenders can be categorised into four main streams according to the employed techniques: tensor factorisation-based, RNN-based, self-attention-based and GNN-based. The first stream of next POI recommenders mainly come from the early studies on user sequential behaviour modelling, which are primarily built upon tensor factorisation, where the relationships among user, POI and time are exploited. By factorising the tensor of POI-POI transition cube, FPMC-LR \cite{ChengYLK13} predicts the user's next move based on the exploited POI transition dependencies with location constraints. STELLER \cite{zhao2016stellar} extends FPMC-LR by additionally introducing a POI-time tensor to capture the user's periodical patterns. PRME \cite{FengLZCCY15} proposes a ranking metric learning framework that learns individual preference and POI sequential dependencies in multiple Euclidean spaces improving the recommendation performance. However, this stream of methods are developed based on the assumption that all the factor components are independent and linearly combined, which neglect modelling of these factor interactions.

With the great success of RNNs in sequential data modelling including NLP \cite{SutskeverMH11,LiuYLZ14,SutskeverVL14} and time-series prediction \cite{ChenYC0WZL18,ChenYCWZL20}, RNNs are widely adopted in the task of next POI recommendation. ST-RNN \cite{LiuWWT16} defines two distance-specific and time-specific projection matrices in RNN for spatio-temporal modelling. Similarly, Time-LSTM \cite{ZhuLLWGLC17} and STGN \cite{ZhaoZLXLZSZ19} enhance the capability of vanilla LSTM model by equipping spatio-temporal gates. LSTPM \cite{SunQCLNY20} exploits the long-term user preferences by identifying affinitive trajectories with a non-local operation. Inspired by the recent advances of self-attention mechanism in sequential recommendation \cite{KangM18}, two state-of-the-art self-attentive models are proposed. GeoSAN \cite{LianWG0C20} discretise the geographical information by a hierarchical gridding module performed on the map with a self-attentive encoder, while STAN \cite{LuoLL21} employs linear interpolation to translate the continuous spatial distance and time information into latent embeddings, and adopts a bi-directional self-attentive network for long-short-term preference encoding. However, self-attentive models make predictions only conditioned on individual trajectories, which ignores the useful transition information from other correlated trajectories. To this end, SGRec \cite{LiCLYH21} and STP-UDGAT \cite{LimHNWGWV20} are proposed, which both construct a global graph from the all observable trajectories, and then employ graph attention network (GAT) to capture the collaborative information across trajectories. Nevertheless, stacking too many GNN layers will result in over-smoothing problem \cite{XuLTSKJ18}, which means GNN models can only attend on close neighbours, while the long-range POI dependencies cannot be fully captured. In contrast, \ssgrec\ takes advantages of both self-attention architecture for long-short-term preference modelling and GNN for collaborative information exploitation by encoding various contextual information from graph signals, while alleviates the mentioned limitations of existing methods.

\subsection{Self-supervised Learning in Recommender Systems}
In recent years, self-supervised learning (SSL) has achieved remarkable success in various areas ranging from NLP \cite{abs-2012-15466,GaoYC21,abs-2005-12766}, graph learning \cite{JiaoXZ0ZZ20,YouCSCWS20,0001XYLWW21} and computer vision \cite{ChenK0H20,He0WXG20}. It constructs implicit supervisory signals from unlabelled data helping the recommenders to learn the subtle patterns from the input data. In light of the promising discriminative ability offered by SSL, it has been an emerging paradigm in recent recommender systems. S$^3$-Rec \cite{ZhouWZZWZWW20} applies random masking on items and item attributes, thus maximising the mutual information over the sequences and item attributes.
Yao et al. \cite{abs-2007-12865} propose to perform feature-level dropout and masking on items and item-category for large-scale item recommendation. SGL \cite{WuWF0CLX21} perturbs the user-item graph structure (i.e., node/edge dropout and random walk) to generate multiple views of the same node, enabling SSL for item recommendation. CLS4Rec \cite{xie2020contrastive} develop random augmentation operators launched on item sequences creating different views of sequence-level representations for SSL. To the best of our knowledge, there is no existing work on exploring the self-supervision signals in next POI recommendation. In \ssgrec, we not only introduce random augmentation operators, which are similar to SGL \cite{WuWF0CLX21} but also devise a global context-aware augmentation operator that explores the POI spatial and temporal correlations from the global view to create high-quality augmented views of the same trajectory graph, allowing the integration of self-supervised learning scheme.
\section{Preliminaries}
In this section, we first formulate the next POI recommendation problem, and then describe the key concepts of the Transformer architecture for sequential data modelling, and give a brief introduction to the mutual information maximisation and self-supervised learning. Finally, we present both sequence-level and global-level graph models, which are designed to capture both the POI transition information from both local and global views over all available check-in trajectories for POI representation learning. We summarise the main notation used in this paper in Table \ref{tab:notation}.

\subsection{Definitions and Notation}
\textit{Definition 1: POI.} A POI represents a certain location on a map, such as a theatre or a park. Let $P=\{p_1, p_2, ..., p_{|P|}\}$ be the set of POIs. Each POI $p \in \sP$ corresponds to a quadruplet, $<ID, \operatorname{Cat}(p), lat, lng>$, which represents its unique ID, POI category label, latitude and longitude, respectively.\\
\textit{Definition 2: Check-in Trajectory.} A check-in trajectory is a sequence of chronologically ordered check-in records. In this paper, the terms ``check-in trajectory'' and ``check-in sequence'' are interchangeable. We use $S_u=\{p_1,p_2,...,p_T\} \in \sS$ to denote a trajectory, where the last check-in $p_T$ is recorded from $u$'s current location, $T$ is the maximum length of all trajectories and $\sS$ is the set of all trajectories. In experiment, if the length of a trajectory is greater than $T$, we only select its most recent $T$ check-ins. Each check-in $p_t \in S_u$ associates to a triplet, $<\operatorname{Cat}({p_t}), \operatorname{time}(p_t), u>$, denoting the visited POI category label, check-in timestamp and user ID, respectively.

\subsection{Problem Formulation}
\textbf{Problem statement}: Next POI Recommendation: Given a trajectory sequence $S_u$ of the user $u$, we aim to recommend top-$K$ POIs that $u$ is most likely to visit at the next time step $T+1$.

\subsection{Self-attention for Sequential Data Modelling}
The self-attention mechanism was first proposed in Transformer \cite{VaswaniSPUJGKP17}, which is a sequential model built by stacking multiple identical Transformer layers, each of which is mainly consisted of two main components: a self-attention module and a position-aware module. Let $\mX=\left[\vx_{1}, \cdots, \vx_{T}\right] \in \mathbb{R}^{T \times d}$ be the input sequence, where $d$ is the dimension size of the input embedding vector. The self-attention module with a scaled dot-product attention function $\operatorname{Attn}(\cdot)$ can be formulated as:
\begin{equation}
    \mQ = \mX\mW_{Q}, \; \mK = \mX\mW_{K}, \; \mV = \mX\mW_{V},
\end{equation}
\begin{equation}
    \operatorname{Attn}(\mX) = \operatorname{softmax}(\frac{\mQ \mK^{\top}}{\sqrt{d}}),
    \label{eq:original_attn}
\end{equation}
where $\mW_{Q} \in \sR^{d \times d}$, $\mW_{K} \in \sR^{d \times d}$, and $\mW_{V} \in \sR^{d \times d}$ are projection spaces that map the input $\mX$ into queries, keys and values, respectively. $\sqrt{d}$ helps to alleviate the gradient vanishing problem of the softmax function. To enhance the capability and stability of Transformer, the multi-head attention mechanism is employed:
\begin{equation}
    \operatorname{MHA}(\mX) = [\operatorname{Attn}_{1}(\mX); \cdots; \operatorname{Attn}_{H}(\mX)]\mW_{O},
\end{equation}
where $\mW_{O} \in \sR^{(H \times d) \times d}$ denotes a transformation matrix, each $\operatorname{Attn_{i}(\cdot)}$ owns a set of projection spaces $\{\mW_{Q}^{i},\mW_{K}^{i},\mW_{V}^{i}\}$, $[\cdot;\cdot]$ denotes a concatenation operation, and $H$ denotes the number of attention heads.

\subsection{Mutual Information Maximisation}
Mutual information (MI) is a measure of two random variables. Specifically, given two random variables $X$ and $Y$, MI measures how much information can be obtained from $Y$ by observing $X$ or vice versa. Formally, the MI between $X$ and $Y$ is:
\begin{equation}
    \operatorname{MI}(X; Y) = \sum_{x \in X, y \in Y} p(x, y) \log (\frac{p(x, y)}{p(x) p(y)}),
\end{equation}
where $p(x)$ and $p(y)$ denote the marginal densities, and $p(x,y)$ is joint density. Suppose $X$ and $Y$ are two different views of an input data (e.g., two different partitions of a trajectory), our goal is to derive a function $f(\cdot)$ that maximises $\operatorname{MI}(X; Y)$ by taking $X=x$ and $Y=y$ as input. However, maximising MI straightforwardly is infeasible in high-dimensional space (i.e., $f(\cdot)$ is a DNN-based encoder). One common practice is to use InfoNCE \cite{YazheInfoNCE} as the loss function to estimate a lower bound on $\operatorname{MI}(X; Y)$:
\begin{equation}
    \mathcal{L}_{InfoNCE} = -\log (\frac{e^{\operatorname{sim}(x, y)}}{\sum_{x_i \in \tilde{X}} e^{\operatorname{sim}(x_{i}, y)}}),
    \label{eq:infonce}
\end{equation}
where $\tilde{X}=\{x_1, x_2, ..., x_K\}$ is a set of $K$ random samples, which have one positive sample drawn from $p(x|y)$ and $K-1$ negative samples obtained from the proposal distribution $p(x)$, and $\operatorname{sim}(\cdot)$ is a similarity function implemented by cosine similarity.

\begin{table}[t]
\caption{Notation. \label{tab:notation}}
\begin{tabularx}{\linewidth}{l|X}
\toprule
Notation&Description\\
\midrule
$\sP$   & the set of all POIs\\
$\sS$   & the set of all trajectories\\
$S_u$   & the historical check-in trajectory of the user $u$: $\{p_1,p_2,...,p_T\}$\\
$d \in \sN$   & latent embedding dimension size\\
$T \in \sN$   & maximum trajectory length\\
$N \in \sN$   & maximum number of neighbours in the global temporal graph\\
$M \in \sN$   & number of bins for the distance bias $b_{\operatorname{dist}(v_{p_i}, v_{p_j})}$\\
$B \in \sN$   & batch size\\
$\alpha \in \sN$   & distance threshold in the global spatial graph\\
$\beta \in \sN$ & POI node dropout rate\\
$\lambda \in \sN$ & a hyperparameter that controls the weight of self-supervised learning task\\
$\gamma \in \sN$ & a hyperparameter that controls the maganitude of $L2$ regularisation\\
$\gG_{s_u}$   & local trajectory graph corresponds to the trajectory $S_u$\\
$\gV_{s_u}, \gE_{s_u}$ & the POI node set and edge set of the local trajectory graph $\gG_{s_u}$\\
$\gG_{gt}, \gG_{gs}$ & global temporal graph and global spatial graph\\
$\gV_{gt}, \gE_{gt}$ & node set and edge set of the global temporal graph\\
$\gV_{gs}, \gE_{gs}$ & node set and edge set of the global spatial graph\\
$\vv_p^{gs} \in \sR^{d}, \vv_p^{gt} \in \sR^{d}$ & pretrained spatial node embedding vector and temporal node embedding vector of POI $p$\\
$\vv_p \in \sR^{d}$ & node embedding vector of $p$\\
$\vz_{\operatorname{deg}^{-}(p)} \in \sR^{d}, \vz_{\operatorname{deg}^{+}(p)} \in \sR^{d}$ & learnable embeddings specified by the incoming degree and outgoing degree of the POI node $v_p \in \gG_{s_u}$\\
$\vz_{\operatorname{pop}(p)} \in \sR^{d}$ & the learnable embedding specified by the number of visits\\
$\mA \in \sR^{(T+1) \times d}$ & position embedding matrix\\
$\operatorname{dist}(v_{p_i}, v_{p_j})$ & geographical distance between POI $p_i$ and $p_j$\\
$\operatorname{spd}(v_{p_i}, v_{p_j})$ & the minimum number of edges travelling from $v_{p_i}$ to $v_{p_j}$ in the graph $\gG_{S_u}$\\
$b_{\operatorname{spd}(v_{p_i}, v_{p_j})}$ & a learnable scalar indexed by $\operatorname{spd}(v_{p_i}, v_{p_j})$\\ $b_{\operatorname{dist}(v_{p_i}, v_{p_j})}$ & a learnable scalar corresponds to the distance $\operatorname{dist}(v_{p_i}, v_{p_j})$\\
$\widehat{\gG}_{S_u}$ & local trajectory graph with a master node added\\
$\mW_s \in \sR^{d \times 2d}$ & the projection matrix for generating the final trajectory representation\\
$\widehat{\vy} \in \sR^{|\sP| \times 1}$ & rankings of all POI candidates for next POI recommendation\\
$\mP \in \sR^{|\sp| \times d}$  & POI embedding matrix\\

$\theta$ & the set of all trainable parameters\\

\bottomrule
\end{tabularx}
\end{table}
\section{Methodology}

In this section, we present the details of our proposed Self-supervised Graph-enhanced POI Recommender, namely S$^2$GRec. Figure \ref{fig:model_overview} illustrates its overall pipeline for next POI recommendation, where the following three four procedures are introduced: 1) local and global graph construction for dual-scale POI correlation modelling; 2) POI representation pretraining that learns spatial and temporal POI embedding vectors for the downstream next POI recommendation task; 3) user preference encoding via graph-enhanced self-attentive layers incorporating spatial and temporal effects from the constructed graphs; 4) self-supervised learning with augmented graphs to further enhance the expressiveness of the learned representations.

\subsection{Local and Global Graph Construction}
In order to capture the complex POI-POI relationships, we propose to simultaneously model the POI dependencies from both \textbf{local} (i.e., a single check-in trajectory) and \textbf{global} (i.e., all check-in sequences) perspectives. To this end, we introduce the construction procedure of the following three types of graphs to facilitate the learning of such dual-scale information, which are generated from the independent check-in sequences (i.e., local trajectory graph), all observed check-in sequences (i.e., global temporal graph), and POI geographical adjacency (i.e., global spatial graph), respectively.

\subsubsection{Local Trajectory Graph Construction}
Learning the temporal patterns (i.e., local patterns) over each consecutive check-in pair within the user's visiting historical sequence is the most critical part in user behaviour understanding \cite{QiuYHC20,Wang0CLMQ20}. To jointly preserve the sequential characteristics, and enable the higher-order item dependency modelling, we convert each check-in trajectory $S_u$ into a directed graph. Specifically, given a check-in sequence $S_u$, we denote its converted local trajectory graph as $\gG_{s_u} = \{\gV_{s_u}, \gE_{s_u}\}$, where we treat each unique visited POI $p \in \sP$ as a graph node $v_p$ in the node set $\gV_{s_u}$ and $\gE_{s_u}$ is a set of directional edges, where each edge $e_{i,j} \in \gE_{s_u}$ represents a pair of consecutive check-ins observed in $S_u$ indicating the user moved from the POI $p_i$ to $p_j$. In addition, to ensure each node still keeps its own information after message propagation, an additional self-loop is added to each node. 

\subsubsection{Global Temporal (GT) Graph Construction}
Most existing next POI recommenders \cite{LiuWWT16,LiSZ18,ZhaoZLXLZSZ19} rely heavily on the POI co-occurrence information within a single check-in sequence, yet fail to explore the complex higher-order movement patterns from other correlated sequences. Thus, we propose to construct a global temporal (GT) graph over all observed check-in sequences, which enables the model to have a better understanding of user preferences from the \textbf{global} view. Particularly, we denote the global temporal graph as an undirected graph $\gG_{gt} = \{\gV_{gt}, \gE_{gt}\}$, where $|\gV_{gt}| = |\gP|$ is the set of all POI nodes containing all observed POIs in $\gP$, and $\gE_{gt}$ indicates the edge set, where each directed edge $e_{gt} \in \gE_{gt}$  corresponds to an observed consecutive visited POI pair from the check-in data. However, some POIs are likely to be unattractive to most users, if there are very few co-occurrence happened between two POIs. To avoid the potential noises when constructing GT-graph, we make each node owns at most $N$ neighbours, which are filtered by the descending order of POI node pair co-occurrence.

\subsubsection{Global Spatial (GS) Graph Construction}
The above mentioned local and global temporal graphs mainly benefit the exploitation of user movement preferences, while the important global spatial factor is not considered. For example, a foodie user who has visited a restaurant is likely to explore another unvisited nearby restaurant in the future. Therefore, we propose to construct a global spatial graph to enable the exploration of unvisited POIs for each user. In particular, we denote an undirected global spatial graph as $\gG_{gs} = \{\gV_{gs}, \gE_{gs}\}$. Here, $\gV_{gs} = \gP$ is the node set containing all POIs, and $\gE_{gs}$ denotes the edge set, where each undirected edge $e_{gs} \in \gE_{gs}$ corresponds to two geographically-adjacent POI nodes. More concretely, two POI nodes $(v^{gt}_{p_i}$ and $v^{gt}_{p_j})$ are connected, if their geographical distance $\operatorname{dist}(v^{gt}_{p_i}, v^{gt}_{p_j})$ is below the threshold $\alpha$. Here, the distance $\operatorname{dist}(v^{gt}_{p_i})$ is measured by the Haversine function, and the distance threshold $\alpha$ is a hyperparameter.

\subsection{Spatial and Temporal POI Representation Pretraining}
To incorporate the global spatial and temporal information into each POI's representation, we employ off-the-shelf graph embedding method, node2vec \cite{GroverL16}, to obtain node embedding vectors from GT-graph and GS-graph, respectively. We denote the node $v_p$'s spatial vector as $\vv_p^{gs} \in \sR^{d}$ and temporal vector as $\vv_p^{gt} \in \sR^{d}$ respectively. We fuse both information together by an element-wise addition operation:
\begin{equation}
    \vv_p = \vv_p^{gs} + \vv_p^{gt}.
\end{equation}
The resulting embedding vector $\vv_p \in \sR^{d}$ will be used to initialise the node features in the following user preference encoder.

\begin{figure}[t]
\includegraphics[width=0.8\textwidth]{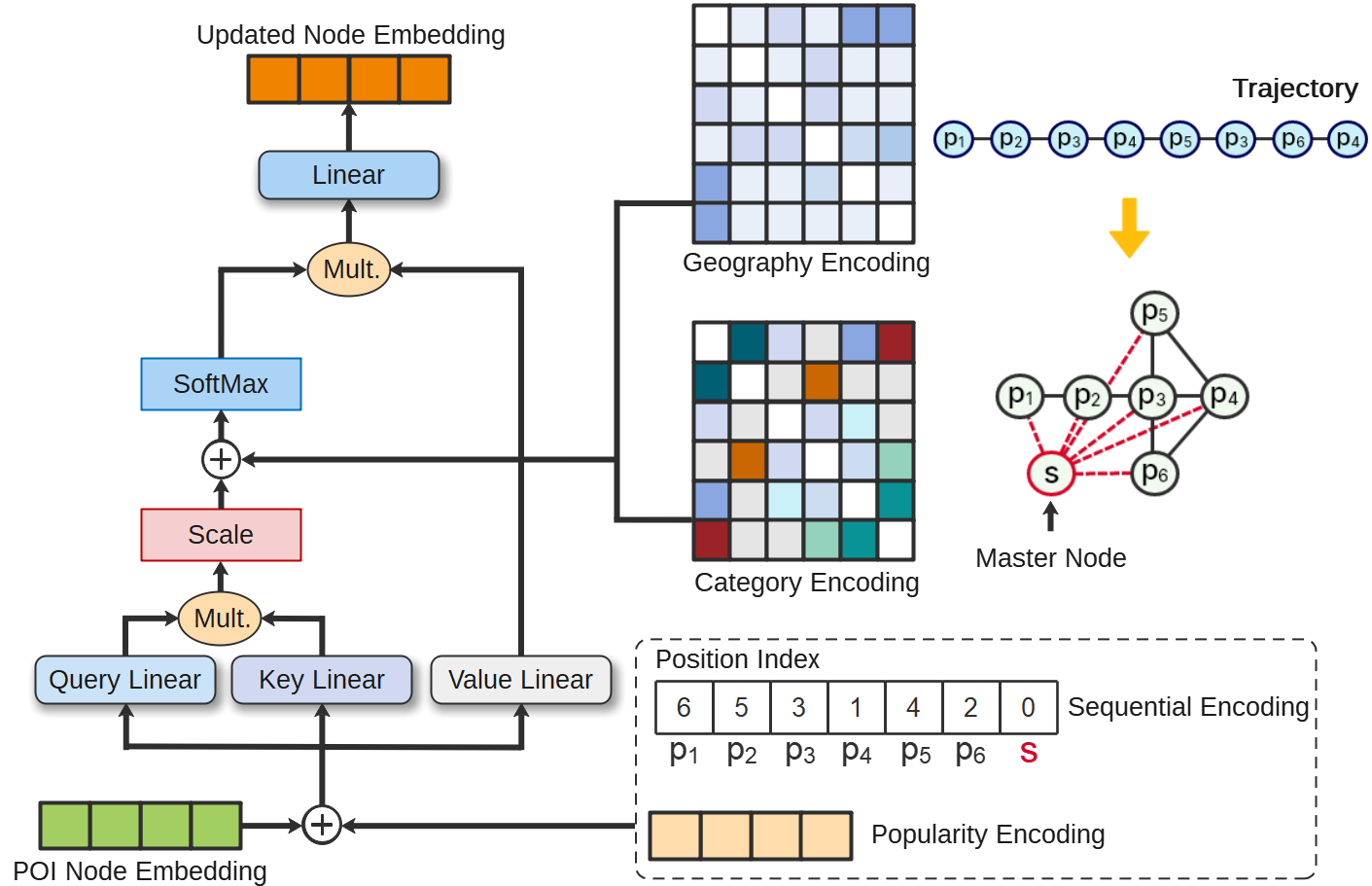}
\caption{An overview of the proposed Graph-enhanced Self-attentive (GSAN) layer.}
\label{fig:GSAN_overview}
\end{figure}

\subsection{Graph-enhanced Self-attentive Network for User Preference Encoding}
The state-of-the-art next POI recommenders, such as GeoSAN \cite{LianWG0C20} and STAN \cite{LuoLL21} are mainly developed based on Transformer architecture \cite{VaswaniSPUJGKP17}, which is only able to explore the movement patterns from independent sequences, while the cross-sequence information is overlooked. Although some recent sequential approaches \cite{QiuHLY20,LiCLYH21} combine both GNN and self-attention mechanism to capture the collaborative information between sequences, their proposed frameworks build two separate GNN and self-attention modules, resulting in model parameter redundancy. Inspired by the recent work on adopting Transformer in graph learning \cite{VijayGT,ChengxuanGraphormer}, we introduce a novel Graph-enhanced Self-attentive Network layer, namely GSAN, which jointly incorporates the collaborative information from the graph and sequential information for the user preference modelling. In particular, as illustrated in Figure \ref{fig:GSAN_overview}, to cope with the the trajectory representation learning from graph data structure using the self-attention architecture, we introduce the following four kinds of encoding: popularity encoding, sequential encoding, geography encoding, and category encoding, allowing the self-attention network to capture the critical sequential patterns as well as leverage the structural information from check-in graphs for higher-order POI interaction modelling.

\subsubsection{Popularity Encoding}
Prior work calculates the node importance distribution based on the semantic correlation among nodes, while the popularity of POIs are ignored. Some studies \cite{YeYLL11,yin2015joint,li2019context} show that the POI popularity is an important indicator for user interest modelling. Thus, instead of relying on attention mechanism for the latent node importance calculation, we propose to inject the global popularity information, which can be regarded as a strong graph structural signal, into each node feature by defining the following three kinds of real-valued embedding vectors based on the constructed global temporal graph:
\begin{equation}
    \tilde{\vh}_p = \vv_p + \vz_{deg^{-}(p)} + \vz_{deg^{+}(p)} + \vz_{pop(p)},
\end{equation}
where $\vz_{\operatorname{deg}^{-}(p)} \in \sR^{d}$, $\vz_{\operatorname{deg}^{+}(p)} \in \sR^{d}$, and $\vz_{\operatorname{pop}(p)} \in \sR^{d}$ are learnable embeddings specified by: (1) incoming node degree $deg^{-}(p)$: the number of unique POIs transiting to $p$; (2) outgoing node degree $deg^{+}(p)$: the number of unique POIs that users have visited travelling from $p$; (3) $pop(p)$: the number of visits, respectively.

\subsubsection{Sequential Encoding}
The next POI recommendation task requires the model to precisely capture the sequential patterns of users. Although converting sequences into graphs can facilitate collaborative information exploration, this also inevitably leads to the loss of partial sequential information due to the nature of graph where the repeatedly-visited POIs are merged into one node. Recent work \cite{LiuZMZ18} suggests that the recent interactions often have greater impacts on current user preference modelling than long-range check-ins. Thus, in this paper, we only record the last occurrence of each POI as the position information in each trajectory. Specifically, we encode each position of check-in in reverse order, such that we assign the index 1 to the last check-in POI $p_T$ and index 2 to the second last check-in $p_{T-1}$, and so on. It is worth noting that the index 0 is a padding index, which will be elaborated in the following section. Thus, given an sample sequence $[p_1, p_2, p_3, p_4, p_5, p_3, p_6, p_4]$, each POI node's corresponding position index is illustrated in Figure \ref{fig:GSAN_overview}. We denote the position embedding matrix as $\mA \in \sR^{(T+1) \times d}$, where $\va_i \in \mQ$ indicates the position embedding at the index $i$, and $T$ is the maximum sequence length. Suppose $p_t \in S_u$ is the check-in at the time step $t$, then by introducing the reverse position-awareness, we can have the position-aware POI node feature $\hat{\vh}_{p_t} \in \sR^{D}$:
\begin{equation}
    \hat{\vh}_{p_t} = \tilde{\vh}_{p_t} + \va_{T-t+1}.
\end{equation}

\subsubsection{Geography Encoding}
\label{sec:spatial_enc}
The key difference between next POI recommendation and sequential recommendation is that the former is supposed to consider the geographical factors when modelling user preferences, since in reality, users are not likely to visit a POI which is far from the current location. To this end, we propose to encode the geographical distance and shortest path distance between each pair of POI nodes in a graph. In particular, given a pair of POI nodes $(v_{p_i}, v_{p_j})$, we first compute their geographical distance using Haversine function represented by $\operatorname{dist}(v_{p_i}, v_{p_j})$, and their shortest-path distance $\operatorname{spd}(v_{p_i}, v_{p_j})$ denoting the minimum number of edges travelling from $v_{p_i}$ to $v_{p_j}$ in the graph $\gG_{S_u}$. Recall that in conventional graph attention networks \cite{VelickovicCCRLB18}, the attention module only performs on each node's first-hop neighbours, which may not be helpful for long-range POI dependency modelling. Considering the local trajectories are usually small, only a limited number of POIs are included in one graph. 
Recall that traditional GNN models only aggregate the first-order neighbours in one layer, which fail to capture long-range dependencies. Therefore, instead of merely calculating the importance of first-order neighbours, our model attends to each pair of POI nodes within the graph $\gG_{S_u}$, which largely increases high-order information propagation. Formally, we modify the original self-attention equation \ref{eq:original_attn} by adding two learnable bias scalars to indicate the two aforementioned distance measurement. Thus, we have the attention score between $v_{p_i}$ and $v_{p_j}$:
\begin{equation}
    \operatorname{Attn}(v_{p_i}, v_{p_j}) = \frac{\mQ \mK^{\top}}{\sqrt{d}} + b_{\operatorname{dist}(v_{p_i}, v_{p_j})} + b_{\operatorname{spd}(v_{p_i}, v_{p_j})},
    \label{eq:attn_spatial}
\end{equation}
where $b_{\operatorname{spd}(v_{p_i}, v_{p_j})}$ is a learnable scalar indexed by $\operatorname{spd}(v_{p_i}, v_{p_j})$, and $b_{\operatorname{dist}(v_{p_i}, v_{p_j})}$ is a learnable scalar produced by linear interpolation. It is worth noting that the geographical distance is a continuous numerical feature which cannot be directly fed into neural models. One common practice is to perform feature discretisation. Inspired by \cite{abs-2004-12602}, we divide the range between minimum distance $min_dist$ and maximum distance $max_dist$ into $M$ bins. We assign learnable scalars to the lower and upper bounds of each bin, thus there are $M+1$ scalars created. If $\operatorname{dist}(v_{p_i}, v_{p_j})$ drops in the $k$-th bin with the lower-bound $lower_k$ and upper-bound $upper_k$, we can perform a linear interpolation to obtain the corresponding bias scalar $b_{\operatorname{dist}(v_{p_i}, v_{p_j})}$:
\begin{equation}
    b_{\operatorname{dist}(v_{p_i}, v_{p_j})} = \frac{b_{upper_k}(upper_k - \operatorname{dist}(v_{p_i}, v_{p_j})) + b_{lower_k}(\operatorname{dist}(v_{p_i}, v_{p_j}) - lower_k)}{upper_k - lower_k}.
\end{equation}

\subsubsection{Category encoding}
The extreme sparsity of POI-POI interactions impedes the recommenders to effectively capture POI-wise sequential dependencies. By contrast, the number of POI category labels is much smaller than that of POI candidates, which leads to denser category-wise interactions than the POI-wise ones. This would be helpful for the model to learn high-level sequential patterns and alleviates the data sparsity issue. Thus, we propose to leverage category-awareness in our GSAN layer by considering category-wise interactions as edge features. Particularly, we define $R$ be a set of undirected category-category pairs, where each $r \in R$ represents a unique category-category pair (e.g., shopping mall-park), and corresponds to a learnable embedding $\vr \in \sR^{d}$. Previous studies of GNNs either add the edge feature to the connected node feature (i.e., first-hop neighbour node feature) \cite{HuFZDRLCL20,abs-2006-07739} or merge the edge feature in the aggregation procedure \cite{GilmerSRVD17,XuHLJ19}. Nevertheless, since our model jointly aggregates both first-order and higher-order neighbour features, how to effectively aggregate the edge features beyond one-hop neighbours has not been investigated in prior work. Thus, we propose a novel category encoding method for graph learning, which can be easily adopted for multi-hop neighbour aggregation. Specifically, assume $v_{p_i}$ and $v_{p_j}$ are high-order neighbours, to estimate the correlations between high-order nodes, we first obtain the shortest path travelling from $v_{p_i}$ and $v_{p_j}$. Denote the shortest path between $v_{p_i}$ and $v_{p_j}$ as $\operatorname{sp}(v_{p_i}, v_{p_j}) = (r_1, r_2, \cdots, r_K)$, which consists of $K$ edges starting from $v_{p_i}$ to $v_{p_j}$. We encode the category information by merging all edge features along the shortest path $\operatorname{sp}(v_{p_i}, v_{p_j})$ to a scalar:
\begin{equation}
    c_{ij} = \frac{1}{K}\sum_{k=1}^{K}\mW_r \vr_k,
\end{equation}
where $\mW_r \in \sR^{d}$ is a trainable weight embedding. To seamlessly incorporate such category information in our attention mechanism (i.e., Equation \ref{eq:attn_spatial}), we treat $c_{ij}$ as a bias term in the attention module. Thus, we can modify Equation \ref{eq:attn_spatial} to the following:
\begin{equation}
    \operatorname{Attn}(v_{p_i}, v_{p_j}) = \frac{\mQ \mK^{\top}}{\sqrt{d}} + b_{\operatorname{dist}(v_{p_i}, v_{p_j})} + b_{\operatorname{spd}(v_{p_i}, v_{p_j})} + c_{ij}.
\end{equation}
Finally, we can update each POI node $v_p \in \gG_{s_u}$ by a attentively-weighted sum over all nodes in $\gG_{s_u}$:
\begin{equation}
    \vv_{p} = \sum_{v_{k} \in \gV_{S_u}}{\operatorname{Attn}(v_{p}, v_{k}) \vv_{k}}.
\end{equation}

\subsubsection{User Preference Encoding}
With our novel GSAN layers, we can obtain each POI node's representation by incorporating sequential, spatial and categorical information. The common practice of existing graph-based sequential recommenders \cite{WuT0WXT19,XuZLSXZFZ19,LiCLYH21} for the sequence-level representation generation from the learned node features is to perform an independent self-attention calculation on the correlations between each item and the last item. However, this results in parameter redundancy as additional attention weights are introduced. Inspired by the recent work of virtual nodes in graph learning \cite{ChengxuanGraphormer,GilmerSRVD17,abs-1709-03741}, we propose to add a master node denoted as $v_s$ into the graph $\gG_{S_u}$, where each node $v_p \in \gV_{S_u}$ connects to $v_s$ (refer to the example illustrated in Figure \ref{fig:GSAN_overview}). The rationale behind this master node is that it not only attentively aggregates each POI node's information based on their contexts for trajectory preference encoding but also facilitates the information propagation among the long-range POIs (i.e., all POI node pairs become either one-hop or two-hop neighbours). Thus, we denote the updated local trajectory graph as $\widehat{\gG}_{S_u} = {\widehat{\gV}_{S_u}, \widehat{\gE}_{S_u}} = {\gV_{S_u} \cup v_s, \gE_{S_u} \cup [(v_1,v_s),(v_2,v_s),\cdots,(v_{|\gV_{S_u}|}, v_s)]}$. We first initialise $v_s$'s corresponding node representation $\vv_s \in \sR^d$ by averaging all the other POI node features:
\begin{equation}
    \vv_s = \frac{1}{|\gV_{S_u}|} \sum_{v_p \in \gV_{S_u}}{\vv_p}.
\end{equation}
Then, we can obtain the trajectory representation $\vv_s$ via the attention:
\begin{equation}
    \vv_s = \sum_{v_p \in \widehat{\gV}_{S_u}}{\operatorname{Attn(v_p, v_s)} \cdot \vv_p}.
\end{equation}
As the decision of next POI recommendation is highly related to the latest check-in \cite{LiCLYH21}, we combine the user's overall preference $\vv_s$ and the last check-in representation $\vv_{p_T}$ to generate the final trajectory representation of the user $u$:
\begin{equation}
    \vs_u = \mW_s [\vv_s;\vv_{p_T}],
\end{equation}
where $\mW_s \in \sR^{d \times 2d}$ is a projection matrix, and $[\cdot;\cdot]$ denotes the concatenation operation.

\subsection{Next POI Prediction}
With the final representation $\vv_s$ that has encoded user spatial and temporal interests, we produce the rankings $\widehat{\vy} \in \sR^{|\sP| \times 1}$ of all POI candidates for next POI recommendation by a dot product operation and normalise it via a softmax function:
\begin{equation}
    \widehat{\vy} = \operatorname{softmax}(\mP \vs_u^{\top}),
\end{equation}
where $\mP \in \sR^{|\sp| \times d}$ is the matrix of all POI node embeddings.
Finally, we apply cross-entropy loss function to quantify the error of next POI prediction task:
\begin{equation}
    \mathcal{L}_{rec} = - \frac{1}{B} \sum_{b=1}^{B} \vy^{\top}_b \log(\widehat{\vy}_b),
\end{equation}
where $b \leq B$ is the index of training samples in a training batch, $\vy_{b}$ is the one-hot vector of next POI ground truth. 

\subsection{Boosting Next POI Recommendation with Self-supervised Learning}
Owing to the exploitation of category-wise spatio-temporal patterns in GSAN layers, the data sparsity issue of POI-wise interactions has been alleviated. However, our model's recommendation capability is still limited by the small amount of observable check-in data. Intuitively, users may not check-in every POI they have visited via the LBSN app, resulting in missing and irregular check-in data. This inevitably has negative impacts on the recommendation performance due to the information loss, which cannot be solved by the category-wise interaction modelling. Besides, so far, our model mainly exploits the local temporal patterns from our local trajectory graph, where the global user preferences are unexplored. To address the above issues and fully explore the rich information from the global view, we innovatively integrate the self-supervised learning (SSL) scheme into the training of \ssgrec.

Previous SSL-based recommendation approaches adopt SSL as either a pretraining task \cite{ZhouWZZWZWW20} or an auxiliary task \cite{abs-2007-12865,0013YYWC021,WuWF0CLX21} to enhance the discrimination ability of recommenders. In this paper, we follow the latter one to supplement the primary next POI prediction task with a self-supervised task. Specifically, we propose the following three augmentation operators performed on local trajectory graphs, which are tailored for the exploration of spatial and sequential correlations from our global graphs: \\
\textbf{POI Node Dropout}: We randomly discard a number of POI nodes by by setting up a drop probability $\beta$ for each node $v_p \in \gV_{S_u}$. It is worth noting that since many local trajectory graphs are small and sparse, hence, if the dropped node $v_p$ makes the graph disconnected, then we establish an edge between any two of $v_p$ first-order neighbour nodes. This augmentation operator stimulates the scenario that some check-in records are missing. \\
\textbf{Correlated Node Insertion}: In practice, the observed check-in sequences are incomplete user behaviours due to the following real cases: 1) Users missed check-ins at some POIs close to the current one (\textbf{spatial}); 2) Users missed check-ins when travelling from one POI to another (\textbf{temporal}).
As a consequence, the model may fail to capture precise user dynamic preferences and neglect comprehensive POI interactions. Motivated by these cases, we propose to insert POI nodes into the trajectory graph $G_{S_u}$ by mining the correlated POI from both GS-graph and GT-graph. To be concrete, we first randomly select $k$ nodes from $\gV_{S_u}$. Then, for each selected node $v_{p_i} \in \gV_{S_u}$, we obtain the most correlated spatial/temporal neighbours $\gN_{s(t)}(v_{p_i})$, where each neighbour POI node $v_{p_j}^{s(t)} \in \gN_{s(t)}(v_{p_i})$ by computing a correlation score between them, which is defined by the cosine similarity function:
\begin{equation}
    \operatorname{correlation}(p_i, p_j) = \frac{\vv_{p_i}^{s(t)} \cdot \vv_{p_j}^{s(t)}}{||\vv_{p_i}||_2||\vv_{p_j}||_2},
\end{equation}
where we use $s(t)$ to indicate spatial (s) and temporal (t) embeddings for simplicity, and $\vv_{p_i}^{s(t)}$ and $\vv_{p_j}^{s(t)}$ are the pretrained embedding vectors of $p_i$ and $p_j$ from global spatial and temporal graphs.\\
\textbf{Correlated Node Substitute}
Motivated by the substitute item recommendation \cite{ChenYYH0020}, which expands user interests by discovering substitutable POIs. This augmentation can be achieved by obtaining the correlated POI $p_j$ using our defined correlation scoring function $\operatorname{correlation}(p_i, p_j)$ for $p_i$. After the node is replaced, we also update the its corresponding edge features.

\subsection{User Preference Encoding via Mutual Information Maximisation}
Having the above proposed augmentation operators, we can employ them to obtain different views of a trajectory graph $\gG_{S_u}$. Denote the encoded trajectory representations of two trajectory graphs augmented from $\gG_{S_u}$ as $\widehat{\vs_u}$ and $\tilde{\vs_u}$, and the representation encode from another trajectory graph $\gG_{S_v}$ as $\tilde{\vs}_v$ by the shared GSAN encoder. We treat each pair of views generated from the same trajectory as a positive pair (i.e., $\{(\widehat{\vs_u}$,$\tilde{\vs}_u) | S_u \in \sS\}$), while the views from different trajectories as a negative pair (i.e., $\{(\widehat{\vs}_u$, $\tilde{\vs}_v) | S_u,S_v \in \sS, S_u \neq S_v\}$). We employ the InfoNCE contrastive loss \cite{GutmannH10} (i.e. Equation \ref{eq:infonce}) to maximise the agreements between positive pairs, while minimise those of negative pairs:
\begin{equation}
    \mathcal{L}_{ssl} = \sum_{S_u \in \sS} -\log (\frac{\operatorname{exp}({\operatorname{sim}(\widehat{\vs}_u,\tilde{\vs}_u)})}{\sum_{S_v \in \sS} \operatorname{exp}({\operatorname{sim}(\widehat{\vs}_u, \tilde{\vs}_v)})}).
\end{equation}

Finally, we combine the recommendation loss and self-supervised loss together via a multi-task learning scheme:
\begin{equation}
    \mathcal{L} = \mathcal{L}_{rec} + \lambda \mathcal{L}_{ssl} + \gamma \|\theta\|^{2}_{2},
\end{equation}
where $\lambda$ is a hyperparameter that decides the weight of the self-supervised task, and $\theta$ is the set of all trainable parameters for $L2$ regularisation under the control of $\gamma$.
\section{Experiments}
\subsection{Experimental Settings}
\subsubsection{Datasets}
We evaluate our proposed \ssgrec\ model on three real-world public check-in datasets: Gowalla \footnote{http://snap.stanford.edu/data/loc-gowalla.html}, Foursquare Tokyo and New York dataset\footnote{https://sites.google.com/site/yangdingqi/home} \cite{LiuLAM13,LiuWSM14}, which have been extensively used in previous studies \cite{LiCLYH21,feng2015personalized,zhao2018go}. In the preprocessing phase, we first discard the inactive users who have less than 10 visiting records and unpopular POIs which are visited by less than 10 users in each dataset. Then, we sort the check-in records of each user in ascending order of timestamp and split them into sequences when the time interval of two consecutive check-ins is longer than 24 hours.

\subsubsection{Task settings \& Evaluation Metrics}
We adopt \textit{leave-one-out} evaluation task to evaluate our models against other state-of-the-art POI recommenders. In each check-in sequence, we take the last check-in (i.e., $p_T$) record as the test sample, the second last check-in (i.e., $p_{T-1}$) as the validation instance, and use the remaining ones (i.e., $[p_1,p_2,...,p_{T-2}]$) for training. We report the performance results on HR$@$\{1, 5, 10, 20\} and nDCG$@$\{1, 5, 10, 20\}, respectively. To avoid potential evaluation biases \cite{KricheneR20}, we rank each ground truth check-in within the whole POI set (i.e., $\sP$) when computing both metrics. For both metrics, higher values indicate better recommendation performance.

\subsection{Implementation Details}
\ssgrec\ is implemented using PyTorch and Torch-geometric framework. We adopt the same dimension size $D=160$ for all embeddings and weight matrices in \ssgrec. The number of heads is set to be 1 and the number of layers is 1. All the parameters are optimised using Adam optimiser with batch size of 32, learning rate of 0.003.

\begin{table*}[!tb]
\caption{The recommendation performance on Gowalla, Foursquare-TKY, and Foursquare-NYC datasets.}
\begin{tabular}{l|l|ccccccc}
\hline
Dataset & Metrics   & FPMC-LR & ST-RNN & ATST-LSTM & LSPTM  & GeoSAN & SGRec  & \ssgrec \\
\hline
Gowalla & HR$@$5    & 0.0373  & 0.1320 & 0.1634    & 0.1798 & 0.1779 & 0.1853 & \textbf{0.2060}                 \\
        & HR$@$10   & 0.0459  & 0.1789 & 0.1990    & 0.2349 & 0.2353 & 0.2394 & \textbf{0.2480}                 \\
        & NDCG$@$5  & 0.0234  & 0.0581 & 0.0894    & 0.1015 & 0.1001 & 0.1067 & \textbf{0.1302}                 \\
        & NDCG$@$10 & 0.0332  & 0.1277 & 0.1605    & 0.1833 & 0.1370 & 0.1839 & \textbf{0.1957}                 \\
\hline
TKY     & HR$@$5    & 0.0845  & 0.1440 & 0.1986    & 0.2402 & 0.3015 & 0.3801 & \textbf{0.4337}                 \\
        & HR$@$10   & 0.1025  & 0.1696 & 0.2840    & 0.3220 & 0.4026 & 0.4757 & \textbf{0.5226}                 \\
        & NDCG$@$5  & 0.0458  & 0.1201 & 0.1650    & 0.1975 & 0.2593 & 0.2836 & \textbf{0.3362}                 \\
        & NDCG$@$10 & 0.0777  & 0.1478 & 0.2191    & 0.2639 & 0.2669 & 0.3050 & \textbf{0.3668}                 \\
\hline
NYC     & HR$@$5    & 0.0477  & 0.1215 & 0.1547    & 0.1867 & 0.2751 & 0.2839 & \textbf{0.2966}                 \\
        & HR$@$10   & 0.0573  & 0.1987 & 0.2329    & 0.2671 & 0.3259 & 0.3333 & \textbf{0.3766}                 \\
        & NDCG$@$5  & 0.0251  & 0.0786 & 0.1012    & 0.1239 & 0.1981 & 0.2007 & \textbf{0.2167}                 \\
        & NDCG$@$10 & 0.0298  & 0.1204 & 0.1357    & 0.1585 & 0.2059 & 0.2204 & \textbf{0.2439}    \\
\hline
\end{tabular}
\label{tab:nextpoi_performance}
\end{table*}

\subsection{Analysis on Recommendation Effectiveness}
We summarise the evaluation results of all models on next POI recommendation task with Table \ref{tab:nextpoi_performance}. From the statistics in the table, we can draw the following observations:
\begin{itemize}
    \item The results on Gowalla, TKY, and NYC datasets show that our proposed SGRec and \ssgrec\ significantly outperform all the other state-of-the-art baseline methods by achieving the second and first places over all metrics, respectively. Compared with the best baseline counterpart, GeoSAN, SGRec and \ssgrec\ respectively gain around 8\% and 15\% performance improvement on average, which confirms the effectiveness of our proposed solution.
    \item The non-deep method, FPMC-LR, shows a clear gap when comparing it to deep methods (i.e., all the other baseline methods and our proposed methods). This suggests that simply using first-order Markov-chain for sequential pattern modelling is insufficient to unveil users' preferences. 
    \item RNN-based methods (i.e., ST-RNN, ATST-LSTM, LSPTM) perform worse than the self-attention based approach (i.e., GeoSAN), which shows the superiority of self-attention in long- and short-term sequential pattern modelling. It is worth mentioning that the LSPTM model that employs a non-local network and a geo-dilated RNN explicitly for long-term and short-term preference encoding, respectively, obtains comparative performance to GeoSAN’s. This further suggests that modelling both long- and short-term user preference is of importance in the success of recommendation in this task.
    \item Our proposed \ssgrec\ consistently outperforms SGRec on all datasets. This is mainly because, by introducing the spatial-temporal aware self-supervised signals during the model training phase, \ssgrec\ learns more discriminative POI and trajectory representations than SGRec, resulting in optimal performance.
\end{itemize}

\subsection{Ablation Study}
\begin{table}[!t]
\caption{Performance of \ssgrec\ variants.}
\centering
\begin{tabular}{|c|c|c|c|c|c|c|}
\hline
\multirow{2}{*}{Method} & \multicolumn{2}{c|}{Gowalla}  & \multicolumn{2}{c|}{TKY} & \multicolumn{2}{c|}{NYC}  \\ \cline{2-7}
    & HR$@$10 & nDCG$@$10 & HR$@$10 & nDCG$@$10 & HR$@$10 & nDCG$@$10 \\ \hline
\ssgrec$_{w/o.pretrain}$ & 0.236 & 0.191 & 0.483 & 0.355 & 0.348 & 0.236 \\
\ssgrec$_{w/o.cate}$     & 0.236 & 0.196 & 0.451 & 0.333 & 0.326 & 0.222 \\
\ssgrec$_{self.attn}$    & 0.228 & 0.194 & 0.507 & 0.361 & 0.366 & 0.241 \\
\ssgrec$_{w/o.ssl}$      & 0.232 & 0.194 & 0.506 & 0.358 & 0.365 & 0.238 \\ \hline
Full Version & 0.248 & 0.196 & 0.523 & 0.367 & 0.377 & 0.244 \\ \hline
\end{tabular}
\label{tab:s2grec_ablation_study}
\end{table}

To better understand the efficacy of different components in \ssgrec, we conducted ablation studies on different degraded versions of \ssgrec. In what follows, we elaborate the details of each degraded variant, and discuss the effectiveness of each component in \ssgrec\ according to the results reported in Table \ref{tab:s2grec_ablation_study}.

\begin{itemize}
    \item \textbf{\ssgrec$_{w/o.pretrain}$}: To validate the impact of the pretraining stage introduced during the global graph construction, we train the POI embeddings from scratch. We can observe that there is a slight performance drop when the embedding pretraining is disabled. This indicates that the pretraining step effectively injects the global contexts into the POI embeddings, improving the expressiveness of the learned representations.
    \item \textbf{\ssgrec$_{w/o.cate}$}: Modelling category-wise transition is one of the most critical component for reducing the negative impacts of data sparsity. We can clearly observe from the table, there is an obvious gap between our full version of \ssgrec and the one without categorical encoding. Since the density of category-level user-POI interactions is naturally higher than the POI-level ones, the ablation test validates the claim that categorical encoding effectively captures the user preferences from denser interactions.
    \item \textbf{\ssgrec$_{self.attn}$}: To evaluate the effectiveness of our designed GSAN layers, we replace them with the conventional self-attention layers \cite{vaswani2017attention}. 
    This version replaces the GSAN layers with the conventional self-attention layers. We can observe that the model experiences a large performance drop when GSAN is removed. This proves that GSAN has stronger ability to capture user preferences and alleviate the sparsity issue than the conventional self-attention mechanism.
    \item \textbf{\ssgrec$_{w/o.ssl}$}: We evaluate the impact of the auxiliary self-supervised objective by removing its loss from the final loss function. From the table, we can see a clear performance degradation when self-supervised loss function is disabled, suggesting the effectiveness of our designed self-supervised learning task for sparsity and noise data.
\end{itemize}

\subsection{Hyperparameter Analysis}

\begin{figure}[th]
 \caption{Impact of (a) dimensionality $D$, (b) self-supervised loss coefficient $\lambda$, (c) node dropout probability $\beta$, and (d) the number of stacked GSAN layers on Gowalla, TKY and NYC datasets.}
 \centering
 \subfloat[][]
 {\includegraphics[width=0.5\columnwidth]{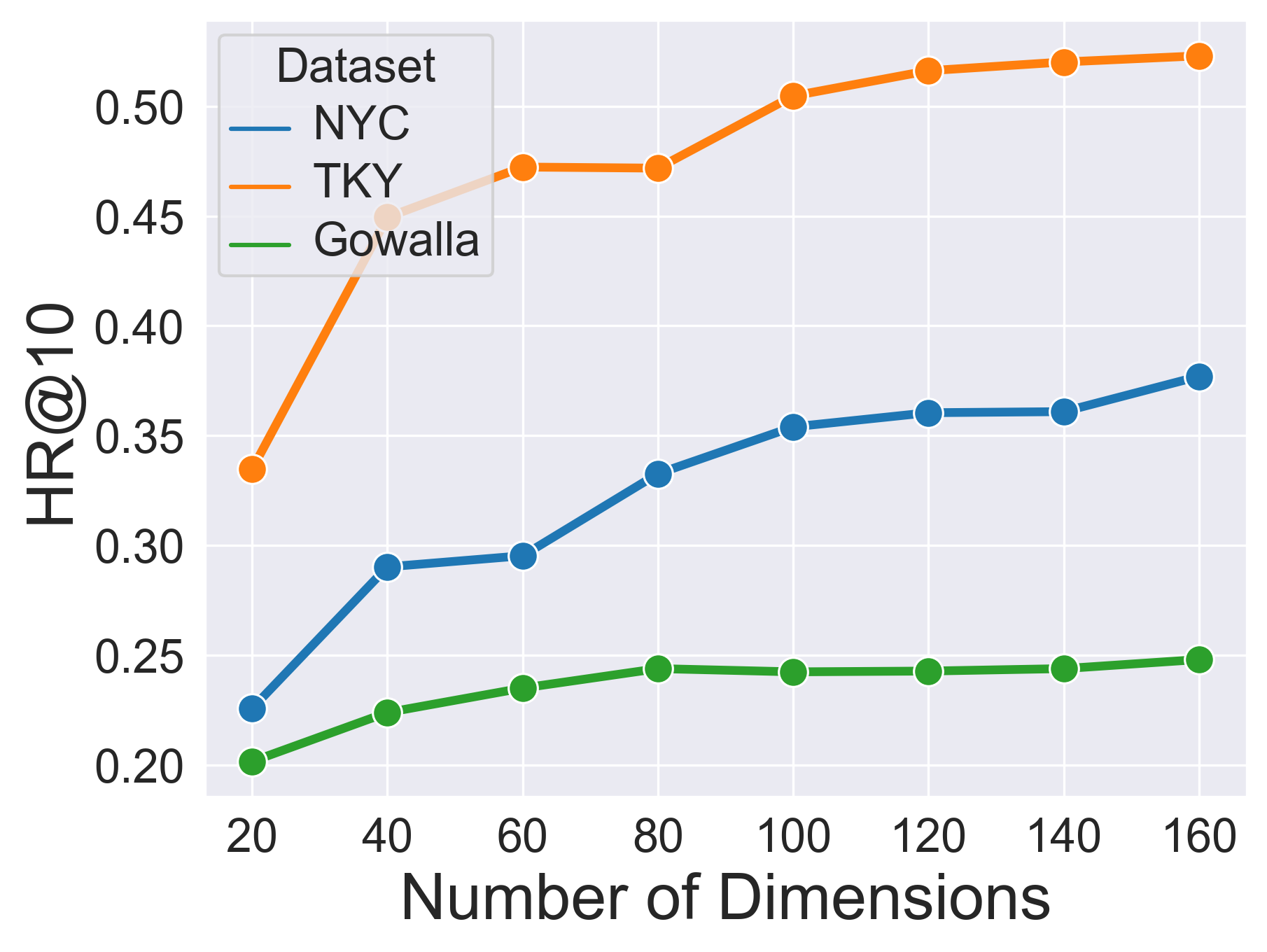}}
 \subfloat[][]
 {\includegraphics[width=0.5\columnwidth]{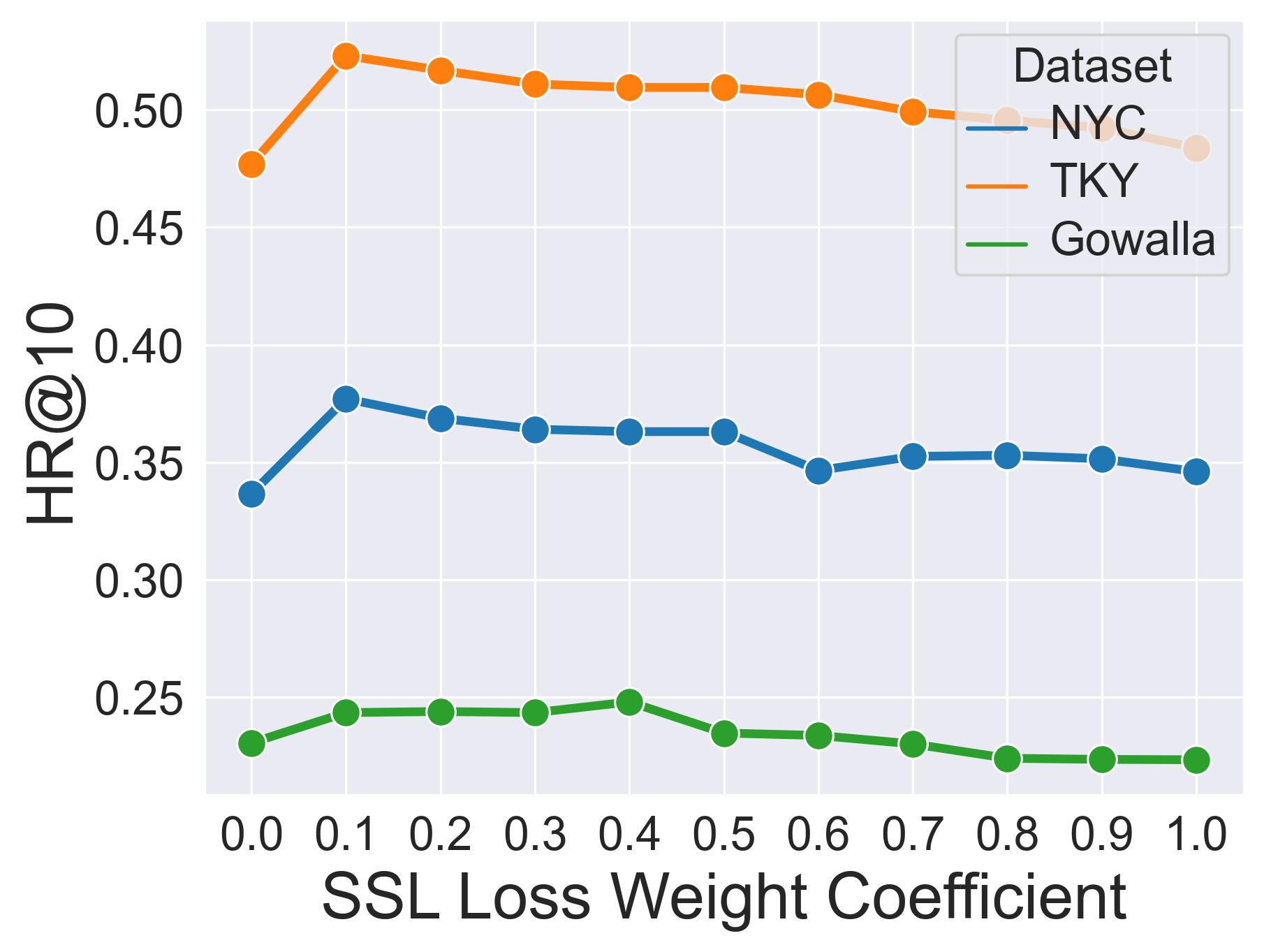}} \\
 \subfloat[][]
 {\includegraphics[width=0.5\columnwidth]{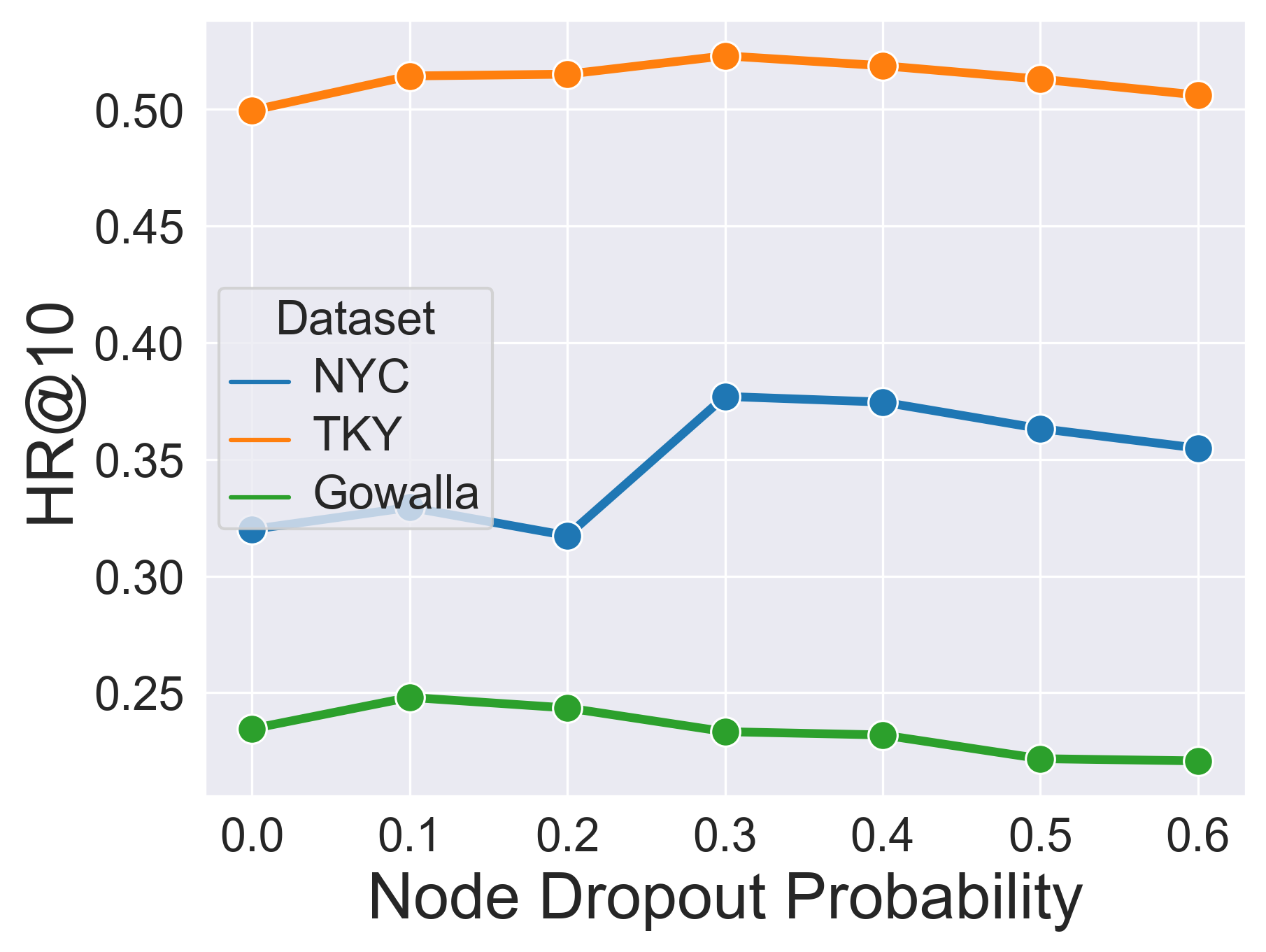}}
 \subfloat[][]
 {\includegraphics[width=0.5\columnwidth]{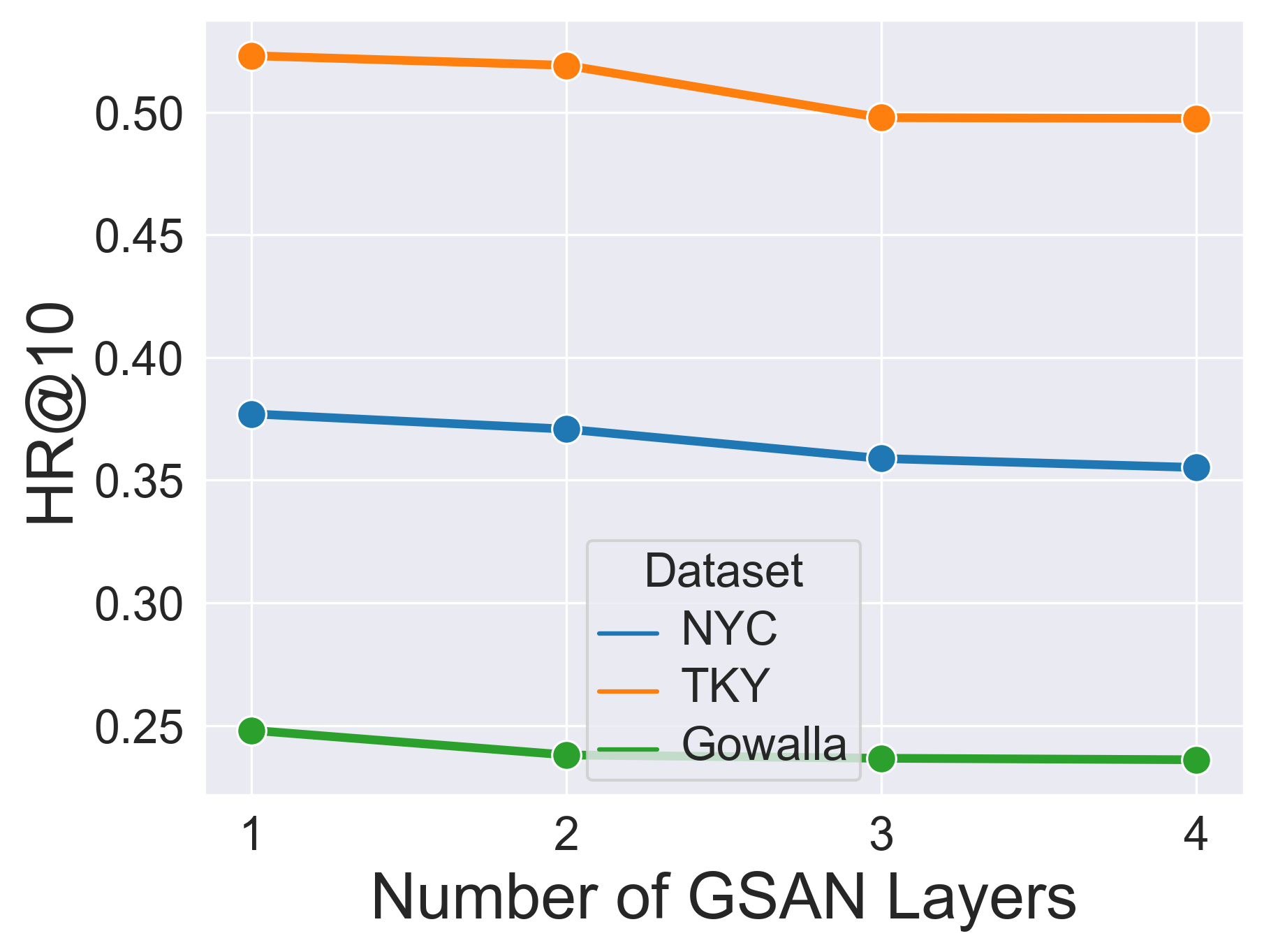}}

 \label{fig:s2grec_hyper_param}
\end{figure}

We also analyse the impact of different hyperparameter settings on \ssgrec, and illustrate the results in Figure \ref{fig:s2grec_hyper_param}. Figure \ref{fig:s2grec_hyper_param} (a) shows the \ssgrec's performance results of various dimensionality. Our model can achieve the best results when $D=160$ on all datasets. Figure \ref{fig:s2grec_hyper_param} (b) shows the model performance for different self-supervised loss coefficients. The performance first increases and quickly reaches a peak, but then falls down, which indicates that the self-supervised objective is beneficial for the model when it does not take over the main task objective. It is worth noting that when comparing with two foursquare datasets where the model has the best performance when $\lambda=0.1$, \ssgrec\'s performance reaches its peak when $\lambda=0.4$ on the Gowalla dataset, which suggests that self-supervised signals play an important role when the data is extremely sparse. Figure \ref{fig:s2grec_hyper_param} (c) illustrates the impact of different dropout probabilities $\beta$. We can observe that the model has best performance when $\beta=0.3$ in the Foursquare datasets, while $\beta=0.1$ in the Gowalla dataset. This is because node dropout increases the sparsity of each training trajectory. If too many nodes are dropped out from the trajectory, our model fails to capture enough transition and preference contexts from the given instance. The last diagram shows the impact of number of GSAN layers. As can be observed from Figure \ref{fig:s2grec_hyper_param} (d), \ssgrec\ obtains best performance when there is only one GSAN layer employed. The performance keeps dropping when the number of layers goes up, which is mainly because too many GSAN layers make the model over-parameterised, and thus, \ssgrec\ can easily become over-fitting.
\section{Conclusion}
To cope with the missing and noisy check-in data issues, we propose a novel solution named \ssgrec\ in this paper. It first obtains graph-enhanced POI representations by incorporating global spatial and temporal information from graph data structures. Then, three node-wise augmentation operators are performed during the model training phase, which boosts the robustness of \ssgrec\ to the data noise and sparsity. Our experiments on three large-scale benchmark datasets show the effectiveness and robustness of \ssgrec, demonstrating the strong feasibility in real-life recommendation scenarios.
\bibliographystyle{ACM-Reference-Format}
\bibliography{main}
\end{document}